\documentclass{article}

\usepackage{PRIMEarxiv}

\usepackage[utf8]{inputenc} %
\usepackage[T1]{fontenc}    %
\usepackage{hyperref}       %
\usepackage{url}            %
\usepackage{booktabs}       %
\usepackage{amsfonts}       %
\usepackage{nicefrac}       %
\usepackage{microtype}      %
\usepackage{lipsum}
\usepackage{fancyhdr}       %
\usepackage{graphicx}       %
\graphicspath{{media/}}     %

\usepackage{amsmath}
\usepackage{tabularx}
\usepackage{todonotes}
\usepackage{multirow}
\usepackage{natbib}[authoryear]
\usepackage{makecell}

\newcommand{\caldist}{calibrated distance}
\newcommand{\Caldist}{Calibrated distance}
\newcommand{\numconcepts}{150}
\newcommand{\numlangs}{20}
\newcommand{\numpairs}{2{,}799}

\title{Convergence in Science, Divergence in Religion:\\
Calibrated Framing Differences Across Wikipedia's Language Editions
}

\author{
  Hung-Hsuan Chen \\
  Computer Science and Information Engineering \\
  National Central University \\
  Taoyuan, Taiwan \\
  \texttt{hhchen1105@acm.org}
}

\begin{document}
\maketitle

\begin{abstract}

When Wikipedia's language editions describe the same concept, how differently do they frame it? Previous work measures coverage gaps between editions; we measure the framing distance for matched concepts. We analyze 2,799 valid articles from 3,000 possible concept-language observations, spanning 150 Wikidata-anchored concepts, 20 language editions, 4 domains, and a calibration set. Raw embedding distances reflect both content differences and how well the encoder aligns each language pair. Even among calibration concepts with stable cross-cultural denotations (e.g., chemical elements, numbers, colors), the largest language-pair mean distance is 3.6 times the smallest, and distances are typically smaller within language families. We define a baseline-adjusted distance (calibrated distance): the distance between two language versions of a concept minus the mean distance for calibration concepts in the same language pair. This adjustment substantially reduces pair-specific alignment differences and the language-family pattern. Across three multilingual encoders (LaBSE, multilingual MPNet, and CMLM), scientific articles align more closely than calibration articles, and all three rank religion first and science/technology last. Concept-level rankings are highly consistent across encoders (Spearman rho=0.75-0.79 for MPNet and CMLM relative to LaBSE). Religion lies significantly above the calibration baseline under LaBSE. Within politics, divergence concentrates on concepts such as censorship and refugee, while democracy and human rights are among the most aligned. Code, data, and per-language-pair calibration baselines are released.\footnote{\url{https://github.com/hhchen1105/cross-linqual-concept}}

\end{abstract}

\section{Introduction}
\label{sec:introduction}

Wikipedia exists in more than 300 languages. Its language editions are largely written independently rather than translated from a common source. When the Turkish, Korean, and Polish editions describe \emph{pilgrimage} or \emph{censorship}, they draw on different editor communities, sources, and cultural contexts. Wikipedia therefore offers a rare corpus for asking: when language communities describe \emph{the same} concept, how differently do they frame it? We find the greatest divergence in religious and politically sensitive concepts, not in settled scientific knowledge.

Previous computational work on cross-lingual Wikipedia differences has focused mainly on \emph{coverage}: which articles, facts, or table entries appear in one edition but not another \citep{samir2024infogap,cappa2025tables}. We instead study \emph{framing}. For concepts matched across editions, does article content diverge more in culturally loaded domains such as religion and politics than in less culturally loaded domains such as basic science and chemical elements? This distinction separates what editions include from how they describe shared topics. It also matters beyond Wikipedia because language editions are core pretraining data for multilingual language models. Such models can change historical narratives with the query language. For example, the same model credits the radio to Popov in Russian but to Marconi in English and Italian; it credits movable-type printing to Bi Sheng in Chinese but to Gutenberg in German \citep{guey2026same}. Measuring divergence in the encyclopedic source material is a first step toward identifying the data-side origins of such language-conditioned behavior.

Multilingual sentence encoders create a methodological problem that previous embedding-based comparisons have not addressed. Encoders such as LaBSE \citep{feng2022labse} align some language pairs much better than others. In our data, the mean distances for the calibration articles range from $0.107$ for Persian--Indonesian to $0.389$ for Hindi--Chinese. These articles cover concepts with relatively stable cross-cultural denotations, including chemical elements, numbers, colors, animals, and natural kinds. Their $3.6\times$ spread therefore reflects encoder alignment and the edition-level writing differences, not only the content. Raw distances are also smaller within language families (e.g., French--Spanish) than across them (e.g., Hindi--Chinese), with a mean of $0.177$ within families versus $0.217$ across them. Without calibration, an analysis could mistake these artifacts for cultural distance. We subtract the mean calibration distance for the same language pair from each raw distance. We call this baseline-adjusted quantity \emph{\caldist}. This correction substantially reduces the aggregate family contrast (see the Results section).

This paper makes three contributions. First, it shows why calibration is necessary: neutral-article baselines vary $3.6\times$ across language pairs and carry a language-family signal. Per-pair calibration substantially weakens both patterns, and parallel text validates the structure of the baselines (see the Method section and the Calibration-set composition subsection). Second, it identifies a clear domain ordering across \numpairs{} articles in \numlangs{} languages spanning more than 12 families and nine scripts. Under LaBSE, religion lies above the calibration baseline ($+0.034$, Holm-adjusted two-sided permutation $p=0.007$), while science lies below it. Political divergence is concentrated in specific concepts: editions describe ``democracy'' similarly but ``censorship'' differently (see the Results section). Third, it separates robust rankings from encoder-dependent magnitudes. Two additional encoders place religion first and science/technology last and reproduce the concept-level ranking (Spearman $\rho = 0.75$--$0.79$), but religion exceeds the baseline significantly only under LaBSE. Excluding bot-created and translation-generated articles changes the religion estimate little (see the Robustness section).

At the concept level, ritual and institutional religious terms (sacrifice, clergy, temple, martyr) and politically sensitive concepts (censorship, refugee) diverge most. Settled knowledge and canonical ideals---evolution, DNA, quantum mechanics, democracy, and human rights---are among the most aligned. These contrasts support a cultural interpretation, but do not identify its cause.

\section{Related Work} \label{sec:related}
\paragraph{Cross-lingual differences between Wikipedia editions.} InfoGap \citep{samir2024infogap} is the closest prior work. It decomposes articles into facts and aligns them across editions with LaBSE and LLM verification. Applied to 2.7K LGBT biographies in English, French, and Russian, it found large coverage gaps and disproportionate inclusion of negative-connotation facts in the Russian edition. InfoGap measures \emph{coverage}: which facts an edition includes. We measure \emph{framing}: how far apart the content of matched articles is after controlling for encoder noise. An edition can present the same facts with different emphasis, or different facts with similar framing, so the measures are complementary. The distinction is not only conceptual: InfoGap's LaBSE-based candidate retrieval is followed by a discrete LLM verification step, so encoder alignment quality mainly affects which candidates are considered rather than the reported fact-presence judgment itself; our continuous distance measure has no such downstream correction and inherits that quality directly, which is why it needs calibration. Extending fact-level alignment to also measure how differently a shared fact is phrased across editions would reintroduce that artifact, making calibration (Eq.~\ref{eq:calibrated-distance}) necessary. InfoGap studies two or three language pairs; our concept-level metric covers all 190 pairs among \numlangs{} languages and compares domains. Other work uses behavioral signals, including overlap in edit-war topics \citep{yasseri2014controversial} and cultural borders inferred from co-editing \citep{samoilenko2016linguistic}. These signals capture editor behavior and topic selection rather than article content. Work on multilingual Wikipedia tables \citep{cappa2025tables} aligns table entries across editions using Wikidata identifiers, and proposes cross-lingual embedding similarity as future work; our calibration would be a necessary addition to that kind of uncalibrated embedding comparison.

\paragraph{Cross-cultural semantics in embedding space.} \citet{sikora2026ssd} compare supervised affective gradients (valence, arousal, dominance) across aligned multilingual embeddings with permutation tests and bootstrap intervals, finding broadly shared valence structure with interpretable residual differences. Their object of study is the affective lexicon; ours is encyclopedic article content. We adopt their permutation-based significance philosophy, adapted to a design where the calibration baseline must be recomputed inside each permutation to avoid construction bias (see the Method section).

\paragraph{Cultural alignment of language models.} A growing line of work probes the sociocultural tendencies of \emph{LLM outputs}: whether they align with survey-based cultural values \citep{arora2023probing,masoud2025cultural}, whether occupational gender patterns in open-ended narratives align more closely with human stereotypes than with labor-force statistics \citep{chen2025stereotypes}, and---closest to our motivation---whether the same model produces different historical attributions depending on the query language \citep{guey2026same}. We ask a data-side question instead, measuring divergence in \emph{human-written Wikipedia content}; the claims therefore do not overlap with the LLM-alignment literature.

\paragraph{Multilingual sentence encoders.} LaBSE \citep{feng2022labse} aligns sentence embeddings across 109 languages through translation-ranking training. Alignment quality varies with the availability of training pairs, contributing to the language-pair variation captured by our baselines. The method itself is encoder-agnostic: substituting another multilingual encoder produces baselines for that encoder and the same edition-level writing differences. We demonstrate this property with two additional encoders \citep{reimers2020making,yang2021universal} in the Encoder robustness subsection.

\section{Data} \label{sec:data}
\subsection{Concepts}
We study \numconcepts{} concepts across four culturally loaded domains---\emph{religion}, \emph{politics}, \emph{science/tech}, and \emph{pop culture}---with 30 concepts each, plus a calibration set of 30 minimally culture-loaded concepts: chemical elements, small numbers, basic colors, common animals, and universal natural kinds, used only to estimate neutral-article baselines (see the Method section). The culturally loaded domains include concepts such as karma, dharma, democracy, liberalism, freedom, human rights, and work-life balance.

Wikipedia titles provide a machine-readable concept vocabulary for language-technology applications, e.g., scientific keyphrase compilation or generation \citep{chen2017keyphrase, chen11collabseer}. We use this vocabulary cross-lingually. Each concept is anchored by a Wikidata QID, which maps it to article titles across editions.

We resolve QIDs from the English article's \texttt{wikibase\_item} page property. A first-hit entity search was unreliable. For example, it mapped \emph{Sin} to Singapore and \emph{Sun} to Sun Microsystems. For the five concepts affected by such disambiguation or redirect errors, we manually verified the correct QID. We recorded it in a documented override file that the resolution pipeline consults instead of the unreliable first-hit search. All \numconcepts{} concepts resolved successfully. The appendix lists them in full.

\subsection{Languages}

We use \numlangs{} languages from 15 families and nine scripts: Arabic, Chinese, English, Finnish, French, German, Hebrew, Hindi, Indonesian, Japanese, Korean, Persian, Polish, Portuguese, Russian, Spanish, Swahili, Thai, Turkish, and Vietnamese. Table~\ref{tab:exclusions} lists their families, scripts, and coverage. Two constraints bound the selection: every language must be supported by LaBSE, and a 20-language sample keeps the 3{,}000-article fetch-and-validation pipeline tractable while providing 190 language pairs. Within that budget, we maximize diversity across families, scripts, and regions.

\subsection{Article text and validation}
For each concept--language observation with a Wikidata sitelink, we retrieve the article's lead section as plain text through the MediaWiki API. Lead sections are more standardized and comparable in length than full articles, reducing length and structure confounding. We exclude observations with no sitelink (i.e., no article exists in that edition), a disambiguation page, or a stub. Stub detection uses a script-aware count. For Chinese, Japanese, and Thai, we count characters rather than whitespace-delimited tokens because these scripts do not consistently separate words with spaces. We consider an extract as a stub if this count is below 20.

Of the $\numconcepts \times \numlangs = 3{,}000$ concept--language observations, \numpairs{} (93.3\%) are valid: 154 lack sitelinks and 47 are stubs, with zero fetch errors. Exclusion is unevenly distributed (Table~\ref{tab:exclusions}): Swahili has 71\% valid observations while every other edition reaches at least 89\%. We treat per-language exclusion as a substantive signal of Wikipedia coverage inequality and report it alongside the distance results.

\begin{table*}[t]
  \centering
  \small
  \caption{The \numlangs{} language editions: family, script, and valid concept--language observations, ascending by coverage. Swahili's low coverage is itself a coverage-inequality finding (the Robustness section shows it is also the most bot- and translation-mediated edition in our sample).}
  \begin{tabular}{llllrr}
\toprule
Code & Language & Family & Script & Valid & Rate \\
\midrule
sw & Swahili & Bantu & Latin & 106/150 & 71\% \\
hi & Hindi & Indo-Aryan & Devanagari & 133/150 & 89\% \\
fi & Finnish & Uralic & Latin & 134/150 & 89\% \\
ko & Korean & Koreanic & Hangul & 136/150 & 91\% \\
pl & Polish & Slavic & Latin & 138/150 & 92\% \\
de & German & Germanic & Latin & 139/150 & 93\% \\
ja & Japanese & Japonic & Japanese & 141/150 & 94\% \\
th & Thai & Kra-Dai & Thai & 141/150 & 94\% \\
id & Indonesian & Austronesian & Latin & 141/150 & 94\% \\
he & Hebrew & Semitic & Hebrew & 142/150 & 95\% \\
ru & Russian & Slavic & Cyrillic & 142/150 & 95\% \\
ar & Arabic & Semitic & Arabic & 143/150 & 95\% \\
pt & Portuguese & Romance & Latin & 143/150 & 95\% \\
vi & Vietnamese & Austroasiatic & Latin & 143/150 & 95\% \\
zh & Chinese & Sinitic & Han & 144/150 & 96\% \\
fr & French & Romance & Latin & 146/150 & 97\% \\
es & Spanish & Romance & Latin & 146/150 & 97\% \\
tr & Turkish & Turkic & Latin & 146/150 & 97\% \\
fa & Persian & Iranian & Arabic & 146/150 & 97\% \\
en & English & Germanic & Latin & 149/150 & 99\% \\
\bottomrule
\end{tabular}

  \label{tab:exclusions}
\end{table*}

\section{Method}
\label{sec:method}

\subsection{Embeddings}
We split each valid lead section into sentences and embed each sentence with LaBSE \citep{feng2022labse}. We L2-normalize each sentence vector so that every sentence contributes equally to the pooled vector regardless of its magnitude, mean-pool the normalized vectors, and L2-normalize the result. This produces one 768-dimensional vector $v_{c,\ell}$ for each concept $c$ and language $\ell$. Embedding computation ran as single-GPU jobs on an internal computing cluster rather than a cloud provider; the full three-encoder pass also completes within a few CPU-hours on a machine without a GPU.

\subsection{Calibrated distance}
Let $d(c,i,j) = 1 - \cos(v_{c,i}, v_{c,j})$, where $\cos(\cdot,\cdot)$ denotes cosine similarity, be the raw distance between the versions of concept $c$ in languages $i$ and $j$. We define

\begin{equation}
e(c, i, j) \;=\; \underbrace{d(c, i, j)}_{\text{raw distance}}
\;-\; \underbrace{\frac{1}{|\mathcal{C}_{ij}|}\sum_{c' \in \mathcal{C}_{ij}} d(c', i, j)}_{\text{calibration floor } b(i,j)}
\label{eq:calibrated-distance}
\end{equation}

Here, $b(i,j) = \frac{1}{|\mathcal{C}_{ij}|}\sum_{c' \in \mathcal{C}_{ij}} d(c', i, j)$ is the calibration floor, where $\mathcal{C}_{ij}$ denotes the pre-selected, minimally culture-loaded calibration concepts (see Data) available in both languages $i$ and $j$ (mean $|\mathcal{C}_{ij}| = 27.5$ out of 30). We call $e(c, i,j)$ the \emph{baseline-adjusted distance}, shortened to \emph{\caldist}. The baseline $b(i,j)$ estimates how far apart languages $i$ and $j$ are on minimally culture-loaded reference concepts. It mainly captures encoder alignment and edition-level writing differences, although some content variation may remain. Baselines range from $0.107$ (fa--id) to $0.389$ (hi--zh); within a language pair, calibration-concept distances have a mean SD of $0.070$, i.e., individual calibration concepts still fluctuate around $b(i,j)$, but this within-pair spread is small relative to the between-pair range, so most of the variation in raw distance is explained by the language pair itself rather than by which calibration concept is used, supporting $b(i,j)$ as a stable per-pair floor. Calibrated distance is signed. A positive value means that two editions describe a concept more differently than they describe the calibration concepts; a negative value means that they describe it more similarly.

\subsection{Significance testing -- reliability of the domain effects}

To establish that a domain's \caldist{} effect is real rather than an artifact of chance, concept selection, or language sampling, we run five checks: a two-sided, concept-level permutation test for whether a domain's mean \caldist{} across language pairs differs from zero; a concept bootstrap for sensitivity to which concepts were chosen; a language-node bootstrap for the sensitivity of concept rankings to which languages are sampled; a concept-node bootstrap for the sensitivity of language rankings to which concepts are sampled; and a Mantel test comparing domain-level geometries.

\paragraph{Permutation test.} Two design choices matter for validity. First, we permute each concept's domain/calibration label rather than treating the 190 language pairs as independent replicates, because every concept contributes distances to all 190 pairs simultaneously, so pairs sharing a concept are correlated and would inflate significance if treated as independent samples. Second, we recompute the calibration baseline $b(i,j)$ inside every permutation rather than holding it fixed, because $b(i,j)$ is itself defined by which concepts are labeled calibration; reusing the original baseline under a shuffled label assignment would bias the null distribution. Concretely, we combine a domain's 30 concepts with the 30 calibration concepts, randomly reassign the domain/calibration label 10{,}000 times, and after each reassignment recompute both $b(i,j)$ and the domain's mean \caldist{} from scratch. Monte Carlo $p$-values use the plus-one correction \citep{north2002note}, which avoids ever reporting $p=0$ under finitely many permutations, followed by Holm adjustment \citep{holm1979simple} to control the family-wise error rate across the four domain tests.

\paragraph{Concept bootstrap.} We compute percentile 95\% intervals from 5{,}000 bootstrap draws (i.e., random sampling with replacement). Each draw resamples the 30 domain concepts and 30 calibration concepts separately, then recomputes the per-pair baseline and domain contrast, applying the same baseline-recomputation principle as the permutation test above. Because the concepts were purposively selected rather than probability-sampled, these intervals measure sensitivity to concept composition; they are not population-sampling intervals for all possible domain concepts.

\paragraph{Language-node bootstrap.} We assess how sensitive the concept-level rankings (the Concept-level structure subsection) are to which languages happen to be sampled. We resample the 20 languages 5{,}000 times, reconstruct the induced dyads, and report each concept's median rank, percentile rank interval, and top- or bottom-10 inclusion rate. This design avoids treating the 190 overlapping dyads as independent, since each language appears in 19 of them -- a different source of non-independence from the concept-sharing addressed above.

\paragraph{Concept-node bootstrap.} Symmetrically, we assess how sensitive the language-level rankings (the Language structure subsection) are to which concepts happen to be sampled. We resample a domain's 30 concepts 5{,}000 times and, on each draw, recompute every language's mean \caldist{} across its \numlangs{}$-1$ pairings and its rank within the domain. We report each language's median rank, percentile rank interval, and top- or bottom-4 inclusion rate, matching the four most divergent and four most aligned languages reported in the Language structure subsection.

\paragraph{Mantel test.} We compare domain-level language-distance matrices with Mantel tests \citep{mantel1967detection}: Spearman correlations of the upper triangles with 10{,}000 label permutations. We apply Holm adjustment \citep{holm1979simple} to the six domain-pair comparisons, following \citet{sikora2026ssd}.

\subsection{Divergence-profile geometry}
Because \caldist{} can be negative, the language-pair matrix is not a valid multidimensional scaling (MDS) dissimilarity matrix. Instead, for each domain we represent each language by a profile: its mean \caldist{} to every language in that domain, with the self-distance set to zero. This makes each profile a \numlangs{}-dimensional vector, where each entry corresponds to one of the \numlangs{} languages and gives the profile language's mean \caldist{} to that target language over the domain's concepts. We center each profile dimension across languages so that the resulting geometry reflects each language's relative pattern of divergence rather than an overall offset on any one target-language dimension. We then compute Euclidean distances between the centered profiles and apply two-dimensional classical MDS. When distances are Euclidean, classical MDS reduces to an eigendecomposition of the double-centered Gram matrix and is equivalent to truncated SVD/PCA on the profile vectors; we use the distance-matrix formulation because it also accepts the non-Euclidean distance definitions used in the sensitivity checks below.\footnote{We use classical MDS rather than nonlinear methods such as t-SNE or UMAP because those methods preserve local neighbor rankings rather than global pairwise distances, offer no analogous stress or variance-explained diagnostic, and are stochastic, which would confound the bootstrap stability check below with optimization randomness rather than data variability.} We report normalized stress and the proportion of positive-eigenvalue variance represented in two dimensions.

To assess stability, we resample the domain's 30 concepts 500 times and rebuild the profiles and MDS. We then correlate pairwise distances in each bootstrap projection with those in the full-sample projection. As sensitivity checks, we also compare the geometry with MDS based on a non-negative constant shift of off-diagonal \caldist{} values and with MDS based on raw cosine distances.

\section{Results}
\label{sec:results}

\subsection{Domain-level divergence}
Table~\ref{tab:domains} shows the domain results under LaBSE. Religion lies significantly above the calibration baseline ($+0.034$, Holm-adjusted $p=.0068$); science/tech lies significantly below it ($-0.024$, Holm-adjusted $p=.030$), meaning its articles align more closely across languages than the calibration articles do. Politics and pop culture do not differ from the baseline. Under LaBSE, culturally loaded content diverges more for religion but not for politics as a whole. Political divergence instead concentrates in specific sensitive concepts (see the Concept-level structure subsection). The Encoder robustness subsection examines which results transfer to other encoders.

\begin{table*}[t]
  \centering
  \small
  \caption{Mean \caldist{} by domain, percentile 95\% concept-bootstrap intervals (5{,}000 draws), and raw and Holm-adjusted two-sided permutation $p$-values (10{,}000 permutations). Domain and calibration concepts are resampled separately, and the per-pair baseline is recomputed on every bootstrap or permutation draw. The 30 concepts per group, not the thousands of overlapping article-pair observations, are the resampling units; because concepts were purposively selected, intervals measure sensitivity to concept composition rather than population sampling uncertainty.}
  \begin{tabular}{lrrrr}
\toprule
Domain & Mean calibrated distance & Concept-bootstrap 95\% interval & Two-sided $p$ & Holm $p$ \\
\midrule
Religion & $+0.0341$ & $[+0.0149,\ +0.0530]$ & $0.002$ & $0.007$ \\
Politics & $+0.0045$ & $[-0.0151,\ +0.0238]$ & $0.653$ & $0.683$ \\
Pop culture & $-0.0085$ & $[-0.0250,\ +0.0079]$ & $0.342$ & $0.683$ \\
Science/tech & $-0.0243$ & $[-0.0416,\ -0.0067]$ & $0.010$ & $0.030$ \\
\bottomrule
\end{tabular}

  \label{tab:domains}
\end{table*}

\subsection{Calibration is necessary}
\label{sec:results-calibration}
Two results show why calibration is necessary. First, neutral-article baselines range from $0.107$ to $0.389$ across language pairs. This $3.6\times$ range is much larger than the domain effects of interest (all absolute mean calibrated distances are below $0.04$), so raw distances largely reflect pair-specific encoder alignment and edition-level writing differences. Second, raw distances average $0.177$ within language families but $0.217$ across families. After calibration, this gap shrinks: mean \caldist{} is $+0.005$ within families and $+0.001$ across them. A smaller script effect remains ($+0.009$ within scripts vs.\ $-0.002$ across scripts), but the domain results also hold within script groups. The Calibration-set composition subsection reports an independent check against content-controlled parallel text, which speaks to whether this shrinkage reflects genuine per-pair encoder alignment rather than an artifact of the embedding pipeline.

\subsection{Language structure} \label{sec:results-langs}

\begin{figure*}[t]
  \centering
  \includegraphics[width=.9\linewidth]{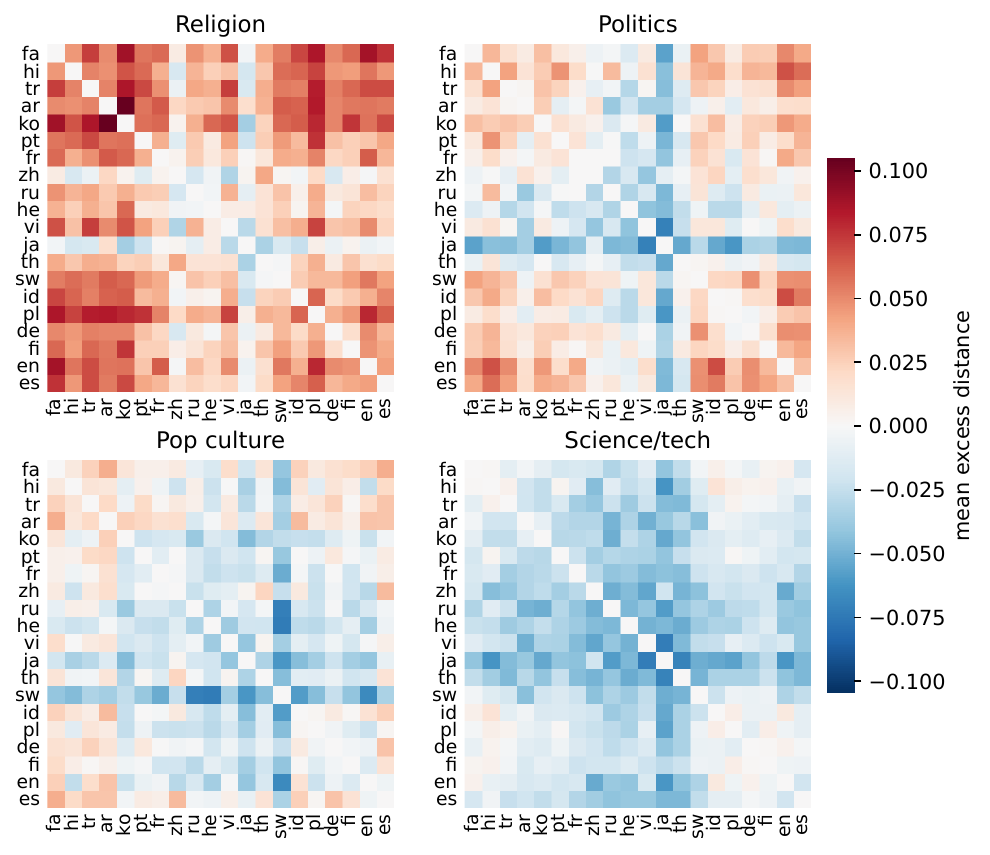}
  \caption{Mean \caldist{} between language editions by domain on a shared scale (red: above the calibration floor; blue: below; gray: no valid pairs). Every panel uses the same language order, obtained by clustering the mean \caldist{} matrix across all four domains. The same cell therefore represents the same language pair in every panel.}
  \label{fig:heatmaps}
\end{figure*}

Figure~\ref{fig:heatmaps} shows the four domain matrices under a shared color scale and a fixed language ordering (hierarchical clustering of the mean \caldist{} matrix over all four domains' concepts). The religion panel is systematically warmer than the science panel, visualizing the domain result. For each language, we average its \caldist{} across all pairings with the other \numlangs{}$-1$ languages in the domain. Within religion, the most divergent languages by this measure are Korean ($+0.056$), Polish ($+0.056$), Persian ($+0.055$), and Turkish ($+0.051$); the most aligned are Japanese ($-0.009$), Chinese ($+0.001$), Hebrew ($+0.018$), and Thai ($+0.021$). Chinese warrants a caveat here: zh.wikipedia has been blocked in the PRC since 2015--2019, so its editor community and this alignment mostly reflect Taiwan, Hong Kong, and overseas Chinese contributors rather than a PRC-based perspective (see the Discussion and Limitations section). The largest single pairs are Arabic--Korean ($0.105$), Persian--Korean ($0.087$), and English--Persian ($0.087$).

Table~\ref{tab:language-rank-stability} assesses how stable this religion-domain ranking is under a concept-node bootstrap (the Method section), which resamples the domain's 30 concepts 5{,}000 times and recomputes each language's mean \caldist{} and rank. The top cluster is more stable in membership than in exact order: Korean, Polish, Persian, and Turkish remain in the top four in 79.9\%, 74.4\%, 81.5\%, and 59.2\% of draws, and Arabic---ranked fifth---is a close contender at 52.0\%. The aligned end is more stable: Japanese and Chinese remain in the bottom four in 100.0\% and 96.3\% of draws, while Hebrew and Thai are less certain (64.8\% and 59.1\%), and Russian---ranked sixteenth---is a comparably close contender at 47.7\%.

\begin{table}[t]
  \centering
  \small
  \caption{Stability of religion-domain language rankings under 5{,}000 concept-node bootstrap draws (resampling the domain's concepts; see the Method section). Rank 1 is most divergent, and rank 20 is most aligned. Tail-4 rate is the proportion of draws in which a language lands in the top or bottom four, shown for languages observed at rank 5 or better, or within 5 of the bottom, since near-threshold languages can also land there under resampling; each language's rate is computed independently, so the shown values do not sum to 100\%.}
  \begin{tabular}{lrrrr}
\toprule
Lang & Mean & Med. rank & 95\% interval & Tail-4 rate \\
\midrule
ko & $+0.056$ & 3 & $[1,\ 8]$ & 79.9\% \\
pl & $+0.056$ & 3 & $[1,\ 10]$ & 74.4\% \\
fa & $+0.055$ & 3 & $[1,\ 7]$ & 81.5\% \\
tr & $+0.052$ & 4 & $[1,\ 10]$ & 59.2\% \\
ar & $+0.051$ & 4 & $[1,\ 16]$ & 52.0\% \\
en & $+0.044$ & 7 & $[3,\ 11]$ & -- \\
hi & $+0.041$ & 8 & $[3,\ 15]$ & -- \\
es & $+0.040$ & 8 & $[3,\ 15]$ & -- \\
pt & $+0.036$ & 10 & $[5,\ 15]$ & -- \\
sw & $+0.036$ & 10 & $[2,\ 17]$ & -- \\
fi & $+0.035$ & 11 & $[4,\ 17]$ & -- \\
fr & $+0.035$ & 11 & $[5,\ 16]$ & -- \\
id & $+0.034$ & 11 & $[6,\ 16]$ & -- \\
vi & $+0.033$ & 12 & $[5,\ 17]$ & -- \\
de & $+0.027$ & 15 & $[10,\ 17]$ & -- \\
ru & $+0.022$ & 16 & $[11,\ 19]$ & 47.7\% \\
th & $+0.020$ & 17 & $[11,\ 19]$ & 59.1\% \\
he & $+0.019$ & 17 & $[11,\ 18]$ & 64.8\% \\
zh & $+0.001$ & 19 & $[16,\ 20]$ & 96.3\% \\
ja & $-0.010$ & 20 & $[19,\ 20]$ & 100.0\% \\
\bottomrule
\end{tabular}

  \label{tab:language-rank-stability}
\end{table}

To check whether religion's language structure reduces to a simple, low-dimensional picture, we project each language's centered divergence profile (the Method section) into two dimensions via classical MDS. We do not visualize the result: the two-dimensional solution captures only 68.6\% of the positive-eigenvalue variance and has normalized stress $0.269$, a moderate-to-poor fit, and individual language positions are unstable under resampling. Across 500 profile-resampling draws, pairwise distances in the bootstrap and full-sample projections have a median Spearman correlation of $0.774$ but a 2.5th percentile of only $0.576$, and specific nearest-neighbor pairings are markedly less stable than the overall geometry (e.g., in an earlier check, one pair of near-neighbor languages selected each other as nearest neighbor in only 33.8\%--51.0\% of bootstrap draws). The overall geometry also correlates only moderately with raw-distance MDS ($\rho=0.47$) and weakly with MDS of constant-shifted calibrated distances ($\rho=0.29$), so even the coarse structure depends on which distance definition is used. Given this fit and sensitivity, we treat the two-dimensional profile geometry as a diagnostic that religion's language structure is not fully summarized by two dimensions, not as a reliable map of which languages pattern together.

\subsection{Concept-level structure}
\label{sec:results-concepts}
Mean \caldist{} produces a clear concept-level pattern. Ritual and institutional religious terms dominate the most divergent concepts: sacrifice ($+0.115$), clergy ($+0.093$), martyr ($+0.085$), temple ($+0.076$), blasphemy ($+0.074$), salvation ($+0.071$), pilgrimage ($+0.066$), and scripture ($+0.064$). The list also includes politically sensitive concepts (censorship $+0.108$, refugee $+0.057$) and locally specific pop-culture terms (streetwear $+0.080$, television drama $+0.063$). By contrast, the most aligned concepts describe standardized bodies of knowledge or canonical ideals: evolution ($-0.072$), democracy ($-0.068$), bacteria ($-0.068$), DNA ($-0.067$), electricity ($-0.066$), civil rights ($-0.064$), and quantum mechanics ($-0.063$).

The extremes are stable under 5{,}000 language-node bootstrap draws: sacrifice and censorship remain in the top ten in 99.0\% and 99.3\% of draws, while concepts near the cutoff are far less stable, e.g., pilgrimage and scripture enter the top ten in only 45.5\% and 37.7\% (Appendix Table~\ref{tab:concept-rank-stability}). We therefore interpret broad tiers rather than an exact ordering. Canonical political concepts also vary: democracy and human rights ($-0.007$) are highly standardized, liberalism is below the floor ($-0.038$), and freedom is near it ($+0.010$). This variation explains why politics averages to the floor even though censorship diverges sharply.

\subsection{Domains share geometry}
All six Mantel correlations between the four domain matrices are positive (Spearman $\rho=0.36$--$0.74$). Four remain significant after Holm correction ($p_{\mathrm{Holm}}\leq.020$). The politics--pop culture and religion--pop culture comparisons are borderline (both $p_{\mathrm{Holm}}=.0502$). Thus, domains share much of their language-pair structure: because Mantel correlations compare each matrix's internal rank order rather than its absolute values, a shared $\rho$ does not require similar raw distances, only a similar relative pattern of which language pairs are more or less divergent than others. Religion's distinctiveness is therefore a shift in overall \emph{level}---its whole matrix runs higher---rather than a wholly different arrangement of which languages pattern together.

\section{Robustness}
\label{sec:robustness}

\subsection{Encoder robustness}
\label{sec:robustness-encoders}
The headline results use LaBSE. To separate data properties from encoder properties, we repeat the entire pipeline with two alternatives. Paraphrase-multilingual-mpnet-base-v2 (MPNet) is a paraphrase-distilled encoder with a 128-token window \citep{reimers2020making}. Universal-sentence-encoder-CMLM (CMLM) shares LaBSE's architecture and 256-token window, but its primary training objective is conditional masked language modeling on monolingual sentence context; its multilingual variant adds bitext retrieval and NLI only as auxiliary co-training tasks, not as the dominant objective the way translation-ranking is for LaBSE \citep{yang2021universal}. For both models, we keep sentence splitting, token windowing, mean pooling, per-pair calibration, and permutation tests unchanged.

Table~\ref{tab:encoder-replication} shows that the ranking replicates. All three encoders place religion first and science/technology last. The concept rankings correlate with LaBSE at Spearman $\rho=0.75$ for MPNet and $0.79$ for CMLM (both $p<10^{-22}$, $n=120$ concepts). The magnitude of the religion effect does not replicate: mean \caldist{} falls from $+0.034$ with LaBSE to $+0.010$ with MPNet and $+0.004$ with CMLM. Neither replication encoder yields a significant domain-level test (Holm-adjusted $p=.37$ and $.76$). Science/technology, by contrast, remains below the baseline under all three encoders (adjusted $p=.030$, $.006$, and $<.001$). LaBSE and CMLM differ in how central cross-lingual alignment is to training: LaBSE optimizes translation-ranking as its sole objective on 6 billion bilingual pairs, while CMLM treats bitext retrieval as an auxiliary task alongside a primarily monolingual objective. This difference in training emphasis and bilingual data scale plausibly contributes to LaBSE's stronger religion effect. We treat the domain and concept ordering as encoder-robust, but the positive religion effect as LaBSE-specific.

\begin{table*}[t]
  \centering
  \small
  \caption{Mean \caldist{} by domain under LaBSE and two replication encoders (MPNet: paraphrase-multilingual-mpnet-base-v2; CMLM: universal-sentence-encoder-CMLM), using the same language-pair-weighted estimand throughout. Each $p$ cell reports the raw two-sided permutation value followed by its Holm-adjusted value in parentheses (10{,}000 permutations; four domains per encoder). The final row gives Spearman correlations of per-concept divergence rankings against LaBSE ($n=120$ concepts). Relative ordering replicates; magnitudes and religion's significance are encoder-dependent.}
  \begin{tabular}{lrrrrrr}
\toprule
 & \multicolumn{2}{c}{LaBSE} & \multicolumn{2}{c}{MPNet} & \multicolumn{2}{c}{CMLM} \\
\cmidrule(lr){2-3}\cmidrule(lr){4-5}\cmidrule(lr){6-7}
Domain & Calibrated & $p$ (Holm) & Calibrated & $p$ (Holm) & Calibrated & $p$ (Holm) \\
\midrule
Religion & $+0.0341$ & $0.002$ ($0.007$) & $+0.0098$ & $0.369$ ($0.369$) & $+0.0043$ & $0.764$ ($0.764$) \\
Politics & $+0.0045$ & $0.653$ ($0.683$) & $-0.0255$ & $0.012$ ($0.035$) & $-0.0423$ & $0.004$ ($0.008$) \\
Pop culture & $-0.0085$ & $0.342$ ($0.683$) & $-0.0191$ & $0.076$ ($0.152$) & $-0.0411$ & $0.003$ ($0.007$) \\
Science/tech & $-0.0243$ & $0.010$ ($0.030$) & $-0.0319$ & $0.001$ ($0.006$) & $-0.0633$ & $<0.001$ ($<0.001$) \\
\midrule
Concept-rank $\rho$ vs.\ LaBSE & \multicolumn{2}{c}{---} & \multicolumn{2}{c}{$0.754$} & \multicolumn{2}{c}{$0.789$} \\
\bottomrule
\end{tabular}

  \label{tab:encoder-replication}
\end{table*}

\subsection{Bot-created and translation-generated articles}
Low-resource editions contain many articles created by bots or with the ContentTranslation tool. These processes may homogenize content and reduce \caldist{}. We audit 889 articles: every valid article in Swahili, Hindi, Thai, Vietnamese, and Indonesian, plus 15 sampled concepts in every other language. We flag an article if it was created through ContentTranslation (identified by the \texttt{contenttranslation} revision tag), created by an account whose username indicates a bot (ends in ``bot''), or has such accounts among more than 50\% of the editors in its last 30 revisions; the MediaWiki revision API we query exposes ContentTranslation tags directly but not an authoritative bot flag, so bot status is a username heuristic rather than a verified account attribute.

Overall, 13.3\% of audited articles are flagged. Swahili has the highest rate at 41.5\% (44/106), followed by Hindi (15\%) and Vietnamese (13\%); most European editions are at or below 7\%. Swahili thus has both the lowest concept coverage and the most bot- and translation-mediated content. Excluding flagged articles changes mean religion \caldist{} only from $+0.034$ to $+0.032$, and no domain changes sign. Because the audit is exhaustive for only five editions and samples 15 concepts elsewhere, it is a subset robustness check rather than a corpus-wide exclusion analysis.

\subsection{Calibration-set composition}
\label{sec:robustness-calibration}
The calibration set contains five subdomains: chemical elements (8 concepts), small numbers (4), basic colors (5), common animals (8), and universal natural kinds such as water and mountain (5). We repeat the analysis five times, omitting one subdomain each time. Mean baselines change by at most $0.011$: from $0.215$ with the full set to $0.205$--$0.223$. Every omission preserves the domain ordering. Religion remains above the baseline, with mean \caldist{} from $+0.026$ (without natural kinds) to $+0.044$ (without colors), and one-sided permutation $p \le .008$ in every case (full set: $+0.034$, $p<.001$). Politics remains at the baseline and science/technology below it; no single calibration subdomain drives the result.

The leave-one-out test cannot detect bias shared by all five subdomains. If every neutral article contains some cultural variation, all Wikipedia-based floors will be inflated. We therefore compare them with floors from FLORES-200 devtest \citep{nllb2022}, where the same 1{,}012 sentences are professionally translated into all \numlangs{} languages. We use the same embedding pipeline and pool ten consecutive sentences into pseudo-documents that approximate the mean calibration-article length. Because the content is parallel, this floor mainly isolates encoder noise; translation from a common English source makes it a lower bound.

Per-pair calibration depends on the \emph{structure} of the floors across language pairs. The Wikipedia and parallel-text floors have similar structure: Spearman $\rho = 0.79$--$0.82$ across pooling levels and Chinese script variants ($n=190$, $p<10^{-41}$). Thus, most pair-to-pair variation removed by our floors reflects encoder alignment. Their average \emph{levels}, however, differ: $0.215$ for Wikipedia articles versus $0.062$ for pooled parallel text. This gap reflects independent authorship---different sentences, emphasis, and style---plus any remaining cultural content in the calibration articles. Both components raise the Wikipedia floor and push \caldist{} downward. Positive values are therefore conservative. Negative science/technology values mean greater alignment than independently written neutral articles, not greater alignment than translation-equivalent text.

\section{Discussion and Limitations}
\label{sec:discussion}

\paragraph{What \caldist{} does and does not measure.} \Caldist{} is relative. It measures how much more differently two editions describe a concept than they describe calibration concepts with stable denotations, under the same encoder. Calibration reduces pair-specific encoder effects and the aggregate language-family contrast, although a small same-script effect remains. \Caldist{} is therefore useful for comparisons across domains, concepts, and language pairs measured in the same way. It is not an absolute or ground-truth measure of cultural difference. It also cannot distinguish two editions that genuinely disagree about a concept's meaning from two editions that emphasize different facets of a concept they would otherwise describe the same way; a high \caldist{} value is consistent with either.

\paragraph{Implications for multilingual language models.} Wikipedia is a major source of pretraining data for multilingual language models. These models can answer the same factual question differently across query languages, such as crediting the radio to Popov or Marconi \citep{guey2026same}. Our results show where such divergence already appears in Wikipedia: in ritual and institutional religious concepts and in specific politically sensitive concepts, not in politics or science as whole domains. Language family alone does not explain the remaining pairwise patterns. \Caldist{} does not directly predict model behavior because pretraining mixtures and post-training also shape outputs. Still, when a model's answers differ by language, our matrices help distinguish disagreement already present in encyclopedic data from disagreement introduced by the model.

\paragraph{Scope and validity limitations.} First, we do not validate the metric against human ratings of semantic difference. Second, language editions are neither nations nor cultures. Chinese illustrates the problem: zh.wikipedia.org is one edition, and its Traditional and Simplified variants display the same underlying content. Because Wikipedia is blocked in the PRC, its editor community is weighted toward Taiwan, Hong Kong, and overseas Chinese communities. Our ``zh'' result therefore reflects a transnational Chinese-language edition, not a PRC-based perspective. Third, we analyze only lead sections; full articles may differ. Fourth, our flags miss manual translations and translations that predate ContentTranslation, so undetected translation may reduce some estimates. Fifth, Swahili has only 71\% concept coverage and a heavily bot-mediated corpus. Sixth, our concept anchoring assumes that a shared Wikidata QID denotes the same conceptual scope across editions; our validation pipeline checks for missing sitelinks, disambiguation pages, and stubs, but does not independently verify scope equivalence, so some divergence for interpretively contested concepts could reflect a scope mismatch rather than differing description of the same referent.

\paragraph{Measurement reliability limitations.} First, rankings replicate across encoders, but the significant positive religion effect appears only under LaBSE (see the Encoder robustness subsection). Absolute magnitudes should not be transferred across encoders. Second, the calibration floor comes from independently written articles, not parallel translations. It therefore absorbs edition style and any remaining cultural content in the neutral concepts, as well as encoder noise. The parallel-text analysis shows that pair-to-pair structure is mainly encoder-driven. Any remaining inflation makes positive \caldist{} conservative, but negative values should be interpreted relative to independently written neutral articles rather than pure encoder noise.

\paragraph{Future work.} Future studies could validate \caldist{} against human ratings, compare independent editions that share a script (e.g., Cantonese Wikipedia), and compare platforms with different editorial governance (e.g., Baidu Baike). Tracking \caldist{} over revision histories could also reveal how framing changes over time.

\section{Conclusion}
\label{sec:conclusion}
We introduced \caldist{}, a cross-lingual metric that subtracts a calibration baseline for each language pair. Across \numconcepts{} concepts, \numlangs{} languages, and four domains plus a calibration set, these baselines vary $3.6\times$ and calibration greatly reduces the language-family contrast. Under LaBSE, religion lies above the baseline, science below it, and political divergence concentrates in specific sensitive concepts rather than the domain as a whole. Two additional encoders preserve the ordering---religion highest and science/technology lowest---but not the significant positive religion effect. Coverage and framing provide complementary views of cross-lingual knowledge inequality. We release code, data, and calibration baselines so future studies can measure framing without mistaking encoder artifacts for culture.

\section*{Acknowledgments and GenAI Usage Disclosure}
We acknowledge support from the National Science and Technology Council of Taiwan under grant number 113-2221-E-008-100-MY3. We thank the National Center for High-performance Computing (NCHC) of National Applied Research Laboratories (NARLabs) in Taiwan for providing computational and storage resources. 

This paper was prepared with substantial assistance from an LLM-based coding agent (Claude Code), used under the author's direction throughout the project: implementing the data and analysis pipeline; proposing and, upon the author's approval, running analyses (e.g., the encoder replications and the parallel-text calibration check); drafting and revising paper text, edited, and in several cases rejected or removed by the author; and locating references, which were verified against their sources before inclusion. Other LLMs (e.g., ChatGPT, Gemini) assisted with language refinement. The author made all research questions, inclusion and exclusion decisions, and final claims, reviewed all content, and takes full responsibility for it.

\bibliographystyle{plainnat}  
\bibliography{ref}

\clearpage
\appendix
\section{Appendix: Concept Inventory}
\label{app:concepts}
Table~\ref{tab:concepts} lists all \numconcepts{} concepts by domain, identified by their English Wikipedia titles; each is anchored by a Wikidata QID in the released data (see the Data section). The calibration set is broken out into its five subdomains, which the leave-one-out analysis of the Calibration-set composition subsection exercises.

\begin{table*}[t]
  \centering
  \small
  \caption{The \numconcepts{} concepts, by domain (English Wikipedia titles).}
  \begin{tabular}{lp{0.72\textwidth}}
\toprule
Domain & Concepts (English Wikipedia titles) \\
\midrule
Religion & Karma, Dharma, Sin, Salvation, Heaven, Hell, Reincarnation, Nirvana, Prayer, Meditation, Fasting, Pilgrimage, Sacrifice, Prophet, Messiah, Monotheism, Polytheism, Atheism, Agnosticism, Secularism, Religious conversion, Religious tolerance, Blasphemy, Martyr, Afterlife, Ritual, Clergy, Temple, Scripture, Creation myth \\
Politics & Democracy, Liberalism, Freedom, Human rights, Socialism, Communism, Conservatism, Nationalism, Populism, Authoritarianism, Revolution, Civil rights, Rule of law, Separation of powers, Political corruption, Propaganda, Censorship, Capitalism, Welfare state, Immigration, Refugee, Colonialism, Imperialism, Self-determination, Patriotism, National security, Freedom of speech, Equality before the law, Political polarization, Work-life balance \\
Science/tech & Artificial intelligence, Machine learning, Internet, World Wide Web, Computer, Algorithm, Database, Cryptography, Quantum mechanics, Relativity, Evolution, Genetics, DNA, Vaccine, Climate change, Nuclear power, Renewable energy, Space exploration, Telescope, Microscope, Electricity, Gravity, Bacteria, Virus, Photosynthesis, Plate tectonics, Periodic table, Semiconductor, Robotics, Biotechnology \\
Pop culture & Anime, Manga, Hip hop music, Rock music, Pop music, Cinema, Hollywood, Bollywood, K-pop, Video game, Esports, Social media, Internet meme, Fandom, Cosplay, Television drama, Reality television, Celebrity, Influencer marketing, Streaming media, Comic book, Superhero, Science fiction, Fantasy, Fashion, Streetwear, Fast food, Coffeehouse, Beauty pageant, Sports fandom \\
Calibration: chemical elements & Hydrogen, Helium, Carbon, Oxygen, Gold, Silver, Iron, Copper \\
Calibration: small numbers & One, Two, Three, Ten \\
Calibration: basic colors & Red, Blue, Green, Black, White \\
Calibration: common animals & Dog, Cat, Horse, Cattle, Chicken, Elephant, Lion, Tiger \\
Calibration: natural kinds & Water, Mountain, River, Sun, Moon \\
\bottomrule
\end{tabular}

  \label{tab:concepts}
\end{table*}

\begin{table*}[t]
  \centering
  \small
  \caption{Stability of the ten highest- and ten lowest-ranked concepts under 5{,}000 language-node bootstrap draws. Rank 1 is most divergent and rank 120 most aligned. Tail identifies the high- or low-divergence block; Tail-10 inclusion is the proportion of draws in which a concept remains in the corresponding top or bottom ten. Wide intervals caution against interpreting adjacent point-estimate ranks as fixed.}
  \begin{tabular}{lllrrrr}
\toprule
Concept & Domain & Tail & Mean & Median rank & Rank 95\% interval & Tail-10 inclusion \\
\midrule
Sacrifice & Religion & High & $+0.115$ & 2 & $[1,\ 8]$ & 99.0\% \\
Censorship & Politics & High & $+0.108$ & 2 & $[1,\ 8]$ & 99.3\% \\
Clergy & Religion & High & $+0.093$ & 4 & $[1,\ 21]$ & 84.1\% \\
Martyr & Religion & High & $+0.085$ & 6 & $[1,\ 19]$ & 81.9\% \\
Streetwear & Pop culture & High & $+0.080$ & 7 & $[2,\ 50]$ & 69.8\% \\
Temple & Religion & High & $+0.076$ & 9 & $[2,\ 29]$ & 60.4\% \\
Blasphemy & Religion & High & $+0.074$ & 9 & $[2,\ 27]$ & 58.6\% \\
Salvation & Religion & High & $+0.071$ & 10 & $[2,\ 33]$ & 53.8\% \\
Pilgrimage & Religion & High & $+0.066$ & 12 & $[2,\ 43]$ & 45.5\% \\
Scripture & Religion & High & $+0.064$ & 13 & $[3,\ 35]$ & 37.7\% \\
\midrule
K-pop & Pop culture & Low & $-0.049$ & 107 & $[84,\ 117]$ & 31.4\% \\
Anime & Pop culture & Low & $-0.060$ & 113 & $[101,\ 118]$ & 67.8\% \\
Quantum mechanics & Science/tech & Low & $-0.063$ & 114 & $[100,\ 120]$ & 70.5\% \\
Civil rights & Politics & Low & $-0.064$ & 114 & $[89,\ 120]$ & 65.2\% \\
Electricity & Science/tech & Low & $-0.066$ & 115 & $[105,\ 120]$ & 81.7\% \\
DNA & Science/tech & Low & $-0.067$ & 115 & $[98,\ 120]$ & 75.6\% \\
Bacteria & Science/tech & Low & $-0.068$ & 116 & $[104,\ 120]$ & 84.9\% \\
Democracy & Politics & Low & $-0.068$ & 116 & $[108,\ 120]$ & 93.5\% \\
Evolution & Science/tech & Low & $-0.072$ & 117 & $[104,\ 120]$ & 89.9\% \\
Virus & Science/tech & Low & $-0.074$ & 118 & $[110,\ 120]$ & 96.8\% \\
\bottomrule
\end{tabular}

  \label{tab:concept-rank-stability}
\end{table*}

\end{document}